\documentclass{article} 
\usepackage{nips11submit_e,times}

\newcommand{\note}[1]{}
\newcommand{\argmin}{\mathop{\arg\min}}

\newcommand{\blue}[1]{{{#1}}}

\usepackage{subfigure}
\usepackage{url}
\usepackage{bm}
\usepackage{graphicx}
\usepackage[boxed, algoruled, vlined, linesnumbered]{algorithm2e} 
\usepackage{times}
\usepackage{dsfont}
\usepackage[longnamesfirst,numbers]{natbib} 
\usepackage{microtype}
\usepackage{amsmath, amssymb, amsfonts}

\makeatletter
\renewcommand\bibsection%
{
  \section*{\refname
    \@mkboth{\MakeUppercase{\refname}}{\MakeUppercase{\refname}}}
}
\makeatother

\makeatletter
\def\url@newstyle{%
  \@ifundefined{selectfont}{\def\UrlFont{\sf}}{\def\UrlFont{\small}}}
\def\url@newFNstyle{%
  \@ifundefined{selectfont}{\def\UrlFont{\sf}}{\def\UrlFont{\scriptsize}}}
\makeatother
\renewcommand{\AlTitleFnt}[1]{{\bf{}#1}}

\newcommand{\hide}[1]{}

\newcommand{\hh}[1]{\note{HH says: #1}}
\renewcommand{\hh}[1]{}

\newcommand{\algofont}[1]{{\footnotesize{\textsc{#1}}}}
\newcommand{\paramils}{\algofont{Param\-ILS}}

\newcommand{\smac}{\algofont{SMAC}}

\newcommand{\spear}{\algofont{SPEAR}}

\newcommand{\cplex}{\algofont{CPLEX}}

\newcommand{\eg}[0]{\emph{e.{}g.{}}}
\newcommand{\ie}[0]{\emph{i.{}e.{}}}

\newcommand{\vTheta}{{\bm{\Theta}}}
\newcommand{\vtheta}{{\bm{\theta}}}

\newcommand{\calM}{\mbox{${\cal M}$}}

\newcommand{\gauss}{\mbox{${\cal N}$}}

\usepackage[outerbars,color]{changebar}

\cbcolor{red}

\SetCommentSty{textit}
\SetKwComment{fhcomment}{// ===== }{}

\SetKwBlock{Procedure}{begin}{end}
\SetKwFunction{Term}{TerminationCriterion()}
\SetKwFunction{Init}{GenerateInitialSolution}
\SetKwFunction{Acc}{AcceptanceCriterion}
\SetKwFunction{Best}{best}
\SetKwFunction{IteratedLocalsearch}{IteratedLocalsearch}
\SetKwFunction{Localsearch}{LocalSearch}
\SetKwFunction{IterativeFirstImprovement}{IterativeFirstImprovement}
\SetKwFunction{Pert}{Perturbation}
\SetFuncSty{textit}

\newtheorem{thm}{Theorem}

\newtheorem{define}[thm]{Definition}
\title{Bayesian Optimization With Censored Response Data}

\author{
Frank Hutter, Holger Hoos, and Kevin Leyton-Brown\\
Department of Computer Science\\
University of British Columbia\\
\texttt{\{hutter, hoos, kevinlb\}@cs.ubc.ca}
}

\nipsfinalcopy 

\begin{document}

\maketitle

\begin{abstract}
Bayesian optimization (BO) aims to minimize a given blackbox function using a model that is updated
whenever new evidence about the function becomes available. Here, we address
the problem of BO under partially \emph{right-censored} response data, where in some evaluations
we only obtain a lower bound on the function value.
The ability to handle such response data allows us to adaptively censor costly function
evaluations in minimization problems where the cost of a function evaluation
corresponds to the function value.
One important application giving rise to such censored data is
the runtime-minimizing variant of the \emph{algorithm configuration} problem:
finding settings of a given parametric algorithm that minimize the runtime
required for solving problem instances from a given distribution.
We demonstrate that terminating slow algorithm runs
prematurely and handling the resulting right-censored observations can substantially
improve the state of the art in model-based algorithm configuration.
\end{abstract}

\section{Introduction}


Right-censored data---data for which only a lower bound on a measurement is available---occurs in several applications. For example, if a patient drops out of a clinical study (for a reason other than death), we know only a lower bound on her survival time.
In some cases, one can actively decide to censor certain data points in order to save time or other resources; for example, if a drug variant $V_1$ is unsuccessful in curing a disease by the time a known drug is successful, one may decide to stop the trial with $V_1$ and instead invest the resources to test a new variant $V_2$.
Here, we describe how to integrate such censored observations into Bayesian optimization (BO).
BO aims to find the minimum of a blackbox function $f: \vTheta \rightarrow \mathds{R}$---a potentially noisy function 
that is not available in closed form, but can be queried at arbitrary input values.
BO proceeds in two phases: (1) constructing a
model of $f$ using the observed function values; and (2) using the model to select the input for the next query.

We extend the standard formulation of blackbox function minimization to include a \emph{cost function} $c:\vTheta \rightarrow \mathds{R}$ that measures the cost of obtaining the function value for a given input.
The budget for minimizing $f$ is now given as a limit on the \emph{cumulative cost} of function evaluations (in contrast to the traditional number of allowed function evaluations).
We call the resulting blackbox function minimization variant $(f,c)$ \emph{cost-varying}.
In this paper, we focus on problems $(f,c)$ with the following \emph{cost monotonicity} property.
\begin{define}
A cost-varying blackbox function minimization problem $(f,c)$ is \emph{cost monotonic} if
\[\forall \vtheta_1, \vtheta_2 \in \vTheta. \left( f(\vtheta_1) < f(\vtheta_2) \Leftrightarrow c(\vtheta_1) < c(\vtheta_2) \right).\]
\end{define}

%
%
For example, a function $f$ may describe how quickly different drug variants cure a disease,
or how quickly plants reach a desired size given different fertilizer variants;
in these examples, it takes exactly time $f(\vtheta)$ to determine $f(\vtheta)$, \ie{}, $c=f$.
When terminating the function evaluation
prematurely after a \emph{censoring threshold} $\kappa < f(\vtheta)$, the cost is only $\kappa$,
but the resulting censored data point is also less informative: we only obtain a lower bound $\kappa<f(\vtheta)$.
Cost monotonicity also applies to minimization objectives other than time, such
as energy consumption,
communication overhead, or strictly monotonic functions of these.

The application domain motivating our research is the following
\emph{algorithm configuration} (AC) problem. We are given a parameterized algorithm $A$, a distribution $D$ of problem instances $\pi \in \mathcal{I}$, and a performance metric $m(\vtheta,\pi)$ capturing $A$'s performance with parameter settings $\vtheta\in \vTheta$ on instances $\pi \in \mathcal{I}$. Let $f(\vtheta) = \mathds{E}_{\pi \sim D} [m(\vtheta,\pi)]$ denote the expected performance of $A$ with setting $\vtheta \in \vTheta$ (where the expectation is over instances $\pi$ drawn from 
$D$; in the case of randomized algorithms, it would also be over random seeds). 
\blue{If we are given only samples $\pi_1, \dots, \pi_N$ from distribution $D$, then $f$ simplifies to $f(\vtheta) = 1/N \cdot \sum_{i=1}^N m(\vtheta,\pi_i)$.}
The problem is then to find a parameter setting $\vtheta$ of $A$ that solves $\argmin_{\theta} f(\vtheta)$.
Automated procedures (\ie{}, algorithms) for solving AC have recently led to substantial
improvements of the state of the art in a wide variety of problem domains, including
SAT-based formal verification~\cite{HutBabHooHu07}, mixed integer
programming~\cite{HutHooLey10-mipconfig}, and automated planning~\cite{FawEtAl11a}.
Traditional AC methods are based on heuristic search~\cite{Minton92minconflict,GraDej92,AdeLag06,ParamILS-JAIR,AnsSelTie09}
and racing algorithms~\cite{BirStuPaqVar02,BirEtAl10}, but recently the BO method \smac{}~\cite{HutHooLey11}
has been shown to compare favourably to these approaches.


A particularly important performance metric $m(\vtheta,\pi)$ in the AC domain
is algorithm A's \emph{runtime} for solving problem instance
$\pi$ given settings $\vtheta$.
Minimizing this metric within a given time budget is a cost monotonic problem (since it yields $c=f$).
Algorithm runs can also be terminated prematurely, yielding a cheaper lower bound on $f$.
%
%
%
As shown in the following, by exploiting this cost monotonicity, we can substantially
improve the state of the art in model-based algorithm configuration.
\section{Regression Models Under Censoring}\label{sec:censored-regression}


Our training data is $(\vtheta_i, y_i, c_i)_{i=1}^n$, where $\vtheta_i \in \vTheta$ is a parameter setting,
$y_i \in \mathds{R}$ is an observation, and $c_i \in \{0,1\}$ is a censoring indicator such that
$f(\vtheta_i) = y_i$ if $c_i=0$ and $f(\vtheta_i) \ge y_i$ if $c_i=1$.
Various types of models can handle such censored data.
In Gaussian processes~(GPs)---the most widely used tool for Bayesian optimization---one could use approximations to handle the resulting
non-Gaussian observation likelihoods; for example, \cite{Ert07} described a Laplace approximation
for handling right-censored data.
Here, we use random forests (RFs)~\cite{Bre01}, which have been shown to yield better
predictive performance for the high-dimensional and predominantly discrete inputs typical of algorithm configuration~\cite{HutHooLey11-EHM}.
Following~\cite{HutHooLey11}, we define the predictive distribution of a RF model $F$ for
input $\vtheta$ as $\gauss(\mu_\vtheta, \sigma^{2}_{\vtheta})$, where $\mu_\vtheta$ and $\sigma^{2}_{\vtheta}$
are the empirical mean and variance of predictions of $f(\vtheta)$ across the trees in $F$.
RFs have previously been adapted to handle censored data~\cite{Seg88,HotEtAl04},
but the classical methods yield non-parametric Kaplan-Meier estimators that do not lend themselves
to Bayesian optimization since they are undefined beyond the largest uncensored data point.


\blue{We introduce a simple EM-type algorithm for filling in censored values.\footnote{\blue{This document is an extended version
of our workshop paper that originally introduced this algorithm~\cite{HutHooLey11-censoring}.}}
We denote the probability density function and the cumulative density function of a
standard Normal distribution by $\varphi$ and $\Phi$, respectively.
Let $\vtheta$ be an input for which we observed a censored value $\kappa < f(\vtheta)$.
Given a Gaussian predictive distribution $\gauss(\mu_\vtheta, \sigma^{2}_{\vtheta})$ of $f(\vtheta)$,
the truncated Gaussian distribution $\gauss(\mu_\vtheta, \sigma^{2}_{\vtheta})_{\ge \kappa}$
is defined by the probability density function
\begin{eqnarray}
\nonumber
p(x) =
\left\{
\begin{array}{lr}
0 & x<\kappa\\
\frac{1}{\sigma_{\vtheta}} \cdot \varphi(\frac{x-\mu_\vtheta}{\sigma_\vtheta}) / (1-\Phi(\frac{\mu_\vtheta-\kappa}{\sigma_\vtheta}))
& x \ge \kappa.\\
\end{array}
\right.
\end{eqnarray}
Our algorithm is inspired by the EM algorithm of Schmee and Hahn~\cite{Josef79}. 
Applied to an RF model as its base model, that algorithm would first fit an initial RF using only uncensored data
and then iterate between the following steps:
\begin{enumerate}
\vspace*{-0.2cm}
	\item[E.] For each tree $\mathcal{T}_b$ in the RF and each $i$ s.t.\ $c_i=1$: $\hat{y}_i^{(b)} \leftarrow \text{ mean of } \gauss(\mu_\vtheta, \sigma^{2}_{\vtheta})_{\ge y_i}$;
\vspace*{-0.2cm}
	\item[M.] Refit the RF using $(\vtheta_i, \hat{y_i}^{(b)}, c_i)_{i=1}^n$ as the basis for tree $\mathcal{T}_b$.
\vspace*{-0.2cm}
\end{enumerate}	
While the mean of 
$\gauss(\mu_\vtheta, \sigma^{2}_{\vtheta})_{\ge \kappa}$
is the best single value to impute,
this algorithm yields overly confident predictions. 
To preserve our uncertainty
about the true value of $f(\vtheta)$, we change Step 1 to:
\begin{enumerate}
\vspace*{-0.2cm}
	\item[E.] For each tree $\mathcal{T}_b$ in the RF and each $i$ s.t.\ $c_i=1$: $\hat{y}_i^{(b)} \leftarrow \text{ sample from } \gauss(\mu_{\vtheta_i}, \sigma^{2}_{{\vtheta_i}})_{\ge y_i}$.
%
\vspace*{-0.2cm}
\end{enumerate}	
}

\blue{More precisely, we draw all required samples for a censored data point at once, using stratifying sampling.
When constructing the random forest, we take bootstrap samples of all $n$ data points for each tree, regardless of their censoring status; this leads to zero, one, or multiple copies of each censored data point per tree. We keep track of the combined number of copies for each data point and obtain these samples as quantiles of the cumulative distribution. Algorithm \ref{alg:Sampling_Schmee_Hahn} summarizes this process in pseudocode. Lines \ref{line:init_start}--\ref{line:end_of_bootstrap} assign a bootstrap sample of the original data to each tree, and line \ref{line:fit_uncensored} initializes the random forest on the uncensored data. Then, the algorithm iterates imputing values for the censored data points (lines \ref{line:impute_start}--\ref{line:impute_end}) and re-fitting trees on both uncensored data points and the individual trees' imputed values for censored data points (lines \ref{line:refit_start}--\ref{line:refit_end}).\footnote{\blue{For software engineering reasons, our actual implementation fits the first random forest using a bootstrap sample different from the one used for the remainder of the iterations.}}}
As an implementation detail, to avoid potentially large outlying predictions above a known maximal runtime $\kappa_{max}$ (in our experiments, $\kappa_{max} = 10\,000$ seconds), we ensure that the mean imputed value does not exceed $\kappa_{max}$.\footnote{In Schmee \& Hahn's algorithm, this simply means imputing $\min \{\kappa_{max}, \text{mean}(\gauss(\mu_i, \sigma^{2}_i)_{\ge z_i})\}$.
In our sampling version, it amounts to keeping track of the mean $m_i$ of the imputed samples for each censored data point $i$ and subtracting $m_i - \kappa_{max}$ from each sample for data point $i$ if $m_i > \kappa_{max}$.} 

Compared to imputing the mean as in a straight-forward adaptation of Schmee \& Hahn's algorithm, our modified version takes our prior uncertainty into account when computing the posterior predictive distribution, thereby avoiding overly confident predictions. 
\blue{We also emulate drawing joint samples for the censored data points (with all imputed values being similarly lucky/unlucky) in order to preserve the predictive uncertainty in the mean of multiple censored values.
In lines \ref{line:impute_start}--\ref{line:impute_end} of Algorithm \ref{line:end_of_bootstrap}, this is done by assigning the lower quantiles of the predictive distribution to trees with lower index (to yield consistent underpredictions) and higher quantiles to the ones with higher index (to yield consistent overpredictions). Using this mechanism preserves our predictive uncertainty even in the mean of $n$ imputed samples, while drawing each sample independently would reduce this uncertainty by a factor of $\sqrt{n}$.}

\NoCaptionOfAlgo
{\DontPrintSemicolon
\small
\begin{algorithm}[tbp]
\setcounter{AlgoLine}{0}
\caption[]{\AlTitleFnt{Algorithm \ref{alg:Sampling_Schmee_Hahn}: Random Forest Fit under Censoring
}}
\SetKwInOut{Input}{Input}
\SetKwInOut{Output}{Output}
\label{alg:Sampling_Schmee_Hahn}
\Input{Data $(\vtheta_i,y_i,c_i)_{i=1}^n$ with $\vtheta_i \in \vTheta$, $y_i \in \mathds{R}$, $c_i \in \{0,1\}$; number of trees, $B$}
\Output{Random forest $\{\mathcal{T}_1, \dots, \mathcal{T}_B\}$}

\lFor(\tcp*[f]{Initialize counter for bootstrap indices}){$j=1, \dots, n$\label{line:init_start}}
{
	$N_j \leftarrow 0$
}

\For(\tcp*[f]{Until line \ref{line:end_of_bootstrap}: draw $n$ bootstrap samples for each of the trees}){$b=1, \dots, B$}
{
	\For{$i=1, \dots, n$}
	{
		$j \leftarrow $ draw integer from $\{1, \dots, n\}$ uniformly at random \tcp*{Resample a data point}
		$(\vtheta_i^{(b)},y_i^{(b)},c_i^{(b)}) \leftarrow (\vtheta_j,y_j,c_j)$ \tcp*{Set the $i$-th data point of tree $b$ to the resampled point}
		$N_j \leftarrow N_j+1$ \tcp*{Total number of times data point j was resampled so far}
		$b_{j,N_j} \leftarrow b$ \tcp*{Tree of this resample of data point j}
		$d_{j,N_j} \leftarrow i$ \label{line:end_of_bootstrap} \tcp*{Index of the resampled data point in that tree}
	}	
}

\lFor(\tcp*[f]{Fit forest using uncensored data}){$b=1, \dots, B$\label{line:fit_uncensored}}
{
	Fit $\mathcal{T}_b$ to data $\{ (\vtheta_i^{(b)}, y_i^{(b)}) \mid c_i^{(b)} = 0\}$
}

\Repeat{converged or iteration limit reached}
{
	\For(\tcp*[f]{Impute samples from truncated predictive distributions\label{line:impute_start}}){$j \in \{k \mid c_k = \text{true}\}$}
	{
		$\gauss(\mu, \sigma^2) \leftarrow $\ predictive distribution of $\{\mathcal{T}_1, \dots, \mathcal{T}_B\}$ for $\vtheta_j$\;			
		$(s_1, \dots, s_{N_j}) \leftarrow q_{1/{(N_j+1)}},\dots,q_{N_j/({N_j+1})}$ quantiles of $\gauss(\mu, \sigma^2)_{\ge y_j}$\hspace*{-2mm} \tcp*{Stratified sampling}
		\lFor{$k=1, \dots, N_j$}
		{
			$\hat{y}_{d_{j,k}}^{(b_{j,k})} \leftarrow s_k$\label{line:impute_end}\;
		}
	}

	\For(\tcp*[f]{Re-fit random forest using uncensored \& imputed data\label{line:refit_start}}){$b=1, \dots, B$}
	{
		fit $\mathcal{T}_b$ to $\{ (x_i^{(b)},y_i^{(b)}) \mid c_i = 0\} \cup \{ (\vtheta_i^{(b)},\hat{y}_i^{(b)}) \mid c_i = 1\}$\label{line:refit_end}\;
	}	
}
\Return $\{\mathcal{T}_1, \dots, \mathcal{T}_B\}$\;
\end{algorithm}
}
\RestoreCaptionOfAlgo

\begin{figure}[tbp]
  \begin{center}
				\subfigure[BO with censoring, initialized with an LHD\label{fig:SMBO-1}]{\includegraphics[width=6.9cm]{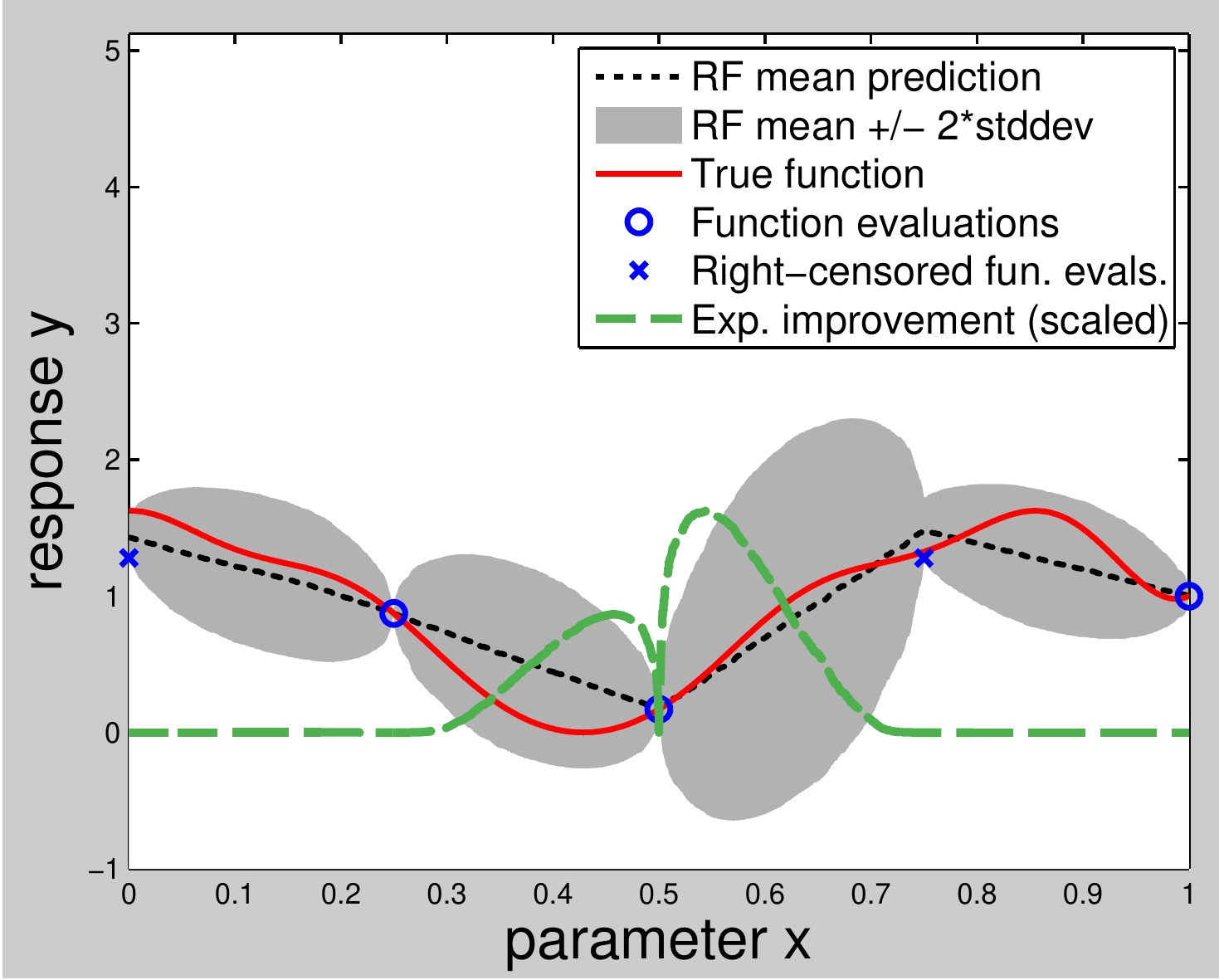}} %
				\subfigure[BO with censoring, step 2\label{fig:SMBO-2}]{\includegraphics[width=6.9cm]{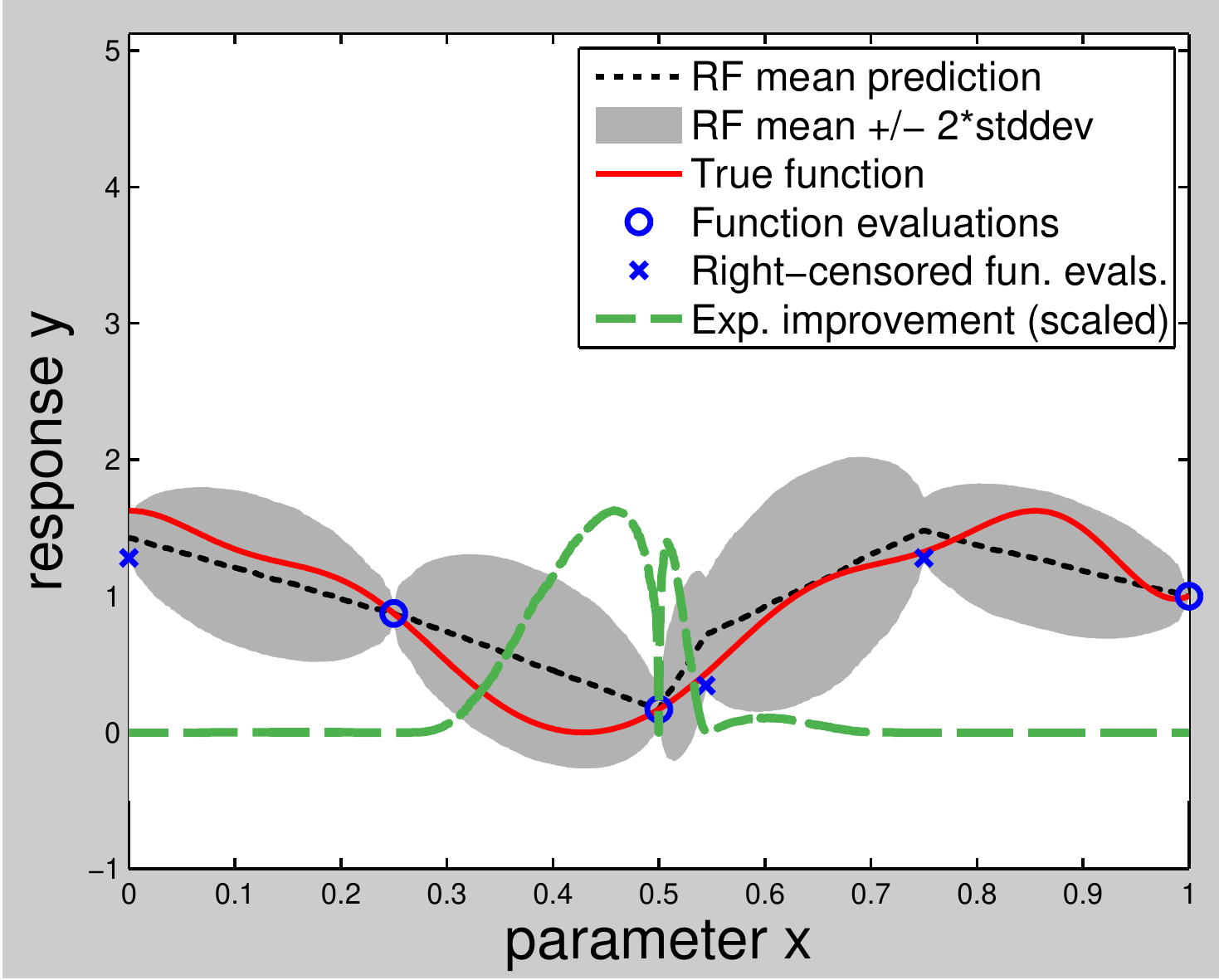}}
				\small
    \caption{\small{}Two steps of Bayesian optimization for minimizing a simple 1-D blackbox functions under censoring, starting from a Latin hypercube design (LHD).
     Circles and x-symbols denote uncensored and right-censored function evaluations, respectively.
     The dotted line denotes the mean prediction of our random forest model with 1000 trees, and the grey area denotes its uncertainty.
     The true function is shown as a solid line and expected improvement (scaled for visualization) as a dashed line.
    \label{fig:smbo-example}}
  \end{center}
\end{figure}

\begin{figure}[tbp]
  \begin{center}
				 \subfigure{\includegraphics[width=6.9cm]{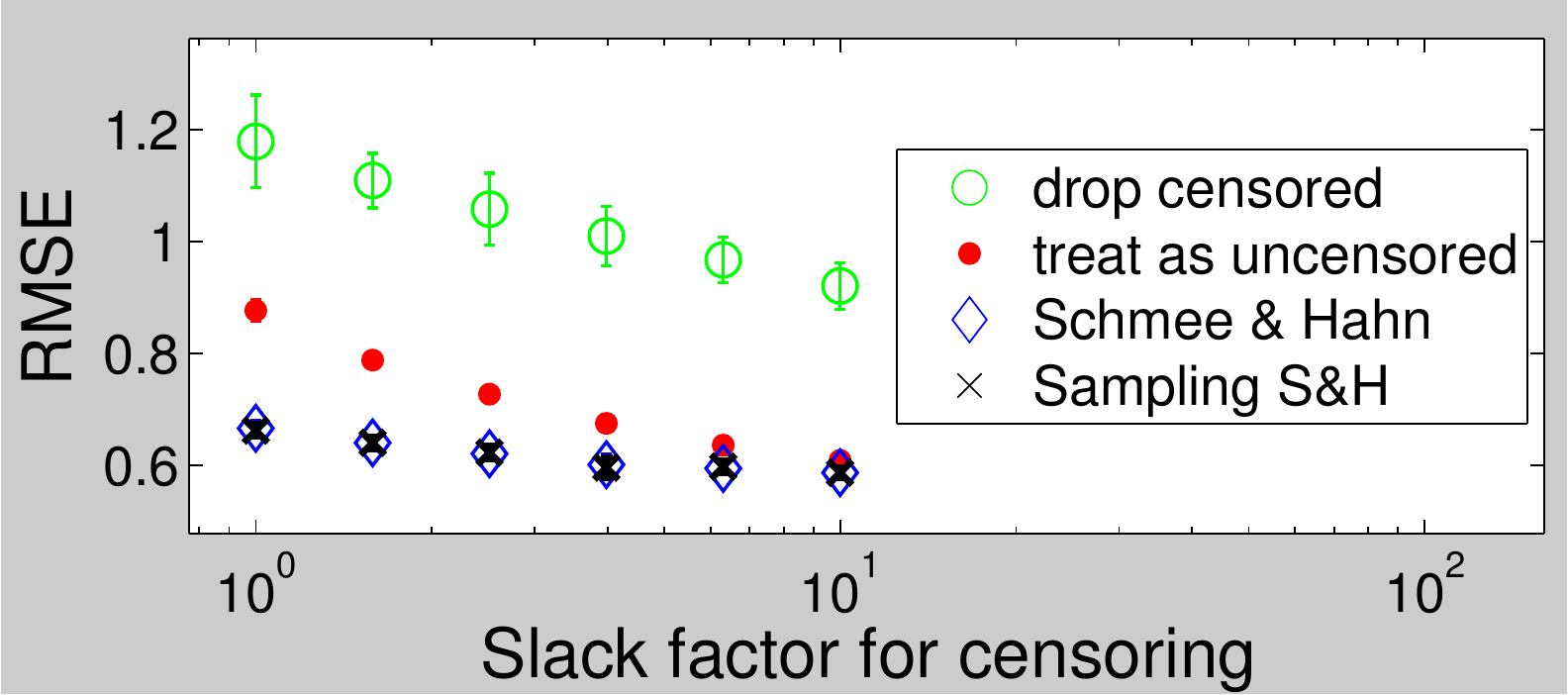}} %
				 \subfigure{\includegraphics[width=6.9cm]{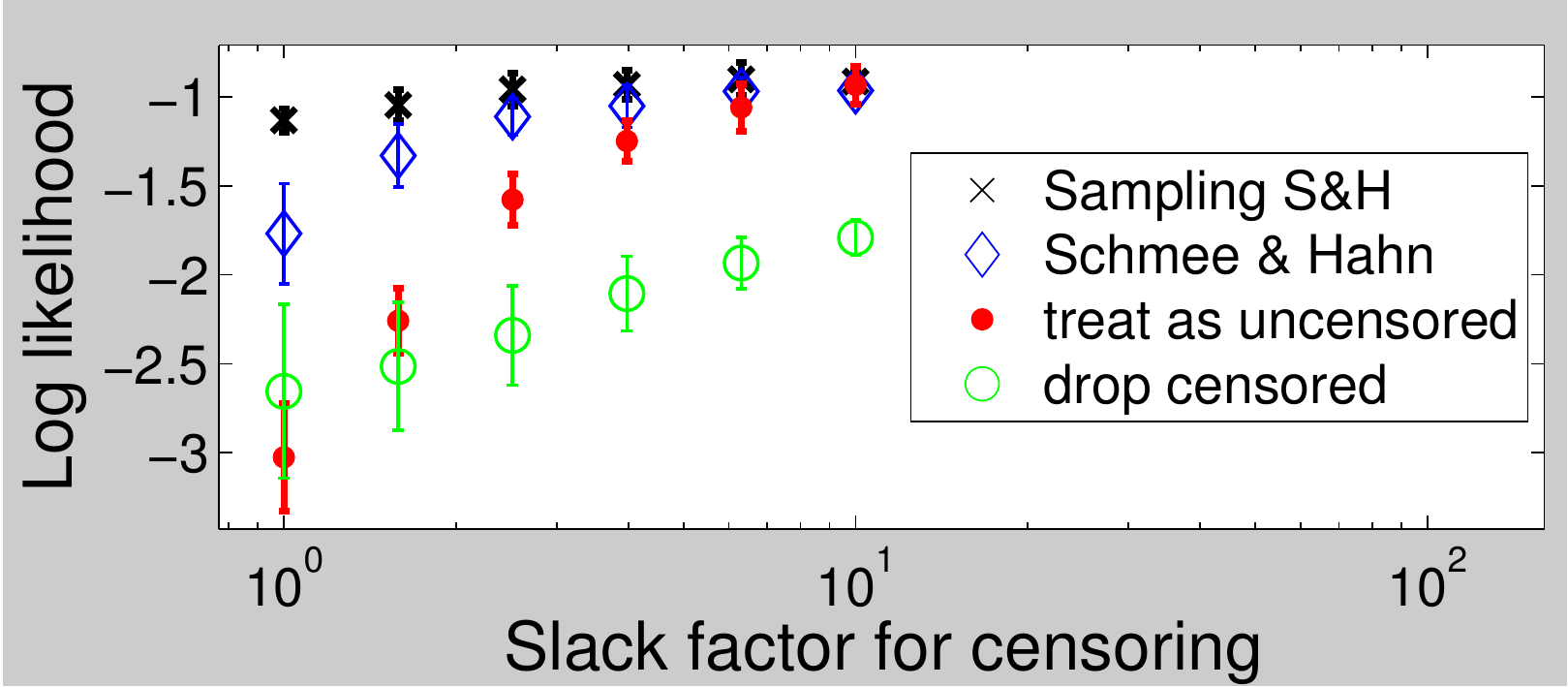}}
				\small
    \caption{\small{}RMSE and log likelihood of various ways of handling censored data with random forests,
    for various strategies of setting the censoring threshold (larger slack factors mean less censoring; see Section \ref{sec:bo-under-cens}).
    \label{fig:model-performance}}
  \end{center}
\end{figure}

Predictive distributions from our sampling-based EM algorithm are visualized for a simple function in Figure \ref{fig:SMBO-1}; note that the predictive variance at censored data points does not collapse to zero (as it would when using Schmee \& Hahn's original procedure with a random forest model).\footnote{Note that our RF's predictive mean converges to a linear interpolation between data points with a sufficient number of trees, and that its variance grows with the distance from observed data points.
(Like the classical approach
for building regression trees, at each node we select an interval $[a,b]$ from which to select a split point to greedily minimize the weighted within-node variance of the node's children. Instead of selecting this point as $(a+b)/2$, we sample it uniformly at random from $[a,b]$. This yields linear interpolation in the limit.)}
As shown in Figure \ref{fig:model-performance}, both Schmee \& Hahn's procedure and our sampling version yield
substantially lower error than either dropping censored data points or treating them as uncensored. By preserving
predictive uncertainty for the censored data points, our sampling method yields the highest log likelihoods.



\section{Bayesian Optimization Under Censoring}\label{sec:bo-under-cens}
\NoCaptionOfAlgo
{
\DontPrintSemicolon
\small
\begin{algorithm}[ht]
\setcounter{AlgoLine}{0}
\caption[]{\AlTitleFnt{Algorithm \ref{alg:EGOcens}: Bayesian Minimization of Blackbox Functions Under Right Censoring}}
\SetKwInOut{Input}{Input}
\SetKwInOut{Output}{Output}
\label{alg:EGOcens}
\Input{Objective function f with input domain $\vTheta = \Theta_1 \times \cdots \times \Theta_d$; budget for optimization}
\Output{Input $\vtheta_{inc} \in \vTheta$ with minimal objective function value found within budget}


Evaluate $f$ at initial design points, yielding data $\{\langle{}\vtheta_i, y_i, c_i\rangle{}\}_{i=1}^{n}$\;
\Repeat{budget for optimization exhausted}
{
Fit regression model $\mathcal{M}$ to the data $\{\langle{}\vtheta_i, y_i, c_i\}_{i=1}^{n}$ collected so far\;
Select $\vtheta_{n+1}$ to maximize an acquisition function defined via $\mathcal{M}$ (\eg{} $EI[\vtheta_{n+1}]$ from Eq. \ref{eqn:exp_imp})\;
Evaluate function at $\vtheta_{n+1}$, yielding $\langle{}y_{n+1}, c_{n+1}\rangle{}$, and increment n\;
}
\end{algorithm}
}
\RestoreCaptionOfAlgo

In order to minimize a blackbox function $f$, Bayesian optimization iteratively evaluates $f$ at some query point, updates a model of $f$, and uses that model to decide which query point to evaluate next. 
One standard method for trading off exploration and exploitation in Bayesian optimization is to
select the next query point $\vtheta$ to maximize the expected positive improvement
$\mathds{E}[I(\vtheta)] = \mathds{E}[\max\{0, f_{\min}-f(\vtheta)\}]$ over the minimal function value $f_{min}$ seen so far.
Let $\mu_{\vtheta}$
and $\sigma_{\vtheta}^2$ denote the mean and variance predicted by our model
for input $\vtheta$, and define
$u = \frac{f_{\min}-\mu_{\vtheta}}{\sigma_{\vtheta}}$.
Then, one can obtain (see, \eg{}, \cite{JonSchWel98})
the closed-form expression
\begin{equation}
\mathds{E}[I(\vtheta)] = \sigma_{\vtheta} \cdot [u \cdot \Phi(u) + \varphi(u)]
\label{eqn:exp_imp},
\end{equation}
where $\varphi$ and $\Phi$ denote the probability density
function and cumulative distribution
function of a standard normal distribution, respectively.
As usual, we maximize this criterion across the input space $\vTheta$ to select the next setting $\vtheta$ to evaluate.
This Bayesian optimization process does not change when we allow some evaluations of $f$ to be right-censored; the only difference is that our model has to be able to handle such censored data. Algorithm \ref{alg:EGOcens} gives the pseudocode of Bayesian optimization for this case of partial right censoring.

In the variant of this problem we face, we can also pick the censoring threshold $\kappa$ for a each query point, up to which we are willing to evaluate the function, and above which we will accept a censored sample. 
There is no obvious best choice of $\kappa$: increasing it yields more informative but also more costly data. Here, we heuristically set $\kappa$ to a multiplicative factor (the ``slack factor'') times $f_{min}$.\footnote{Our approach here is inspired by the ``adaptive capping'' method used by the algorithm configuration procedure \paramils{}~\cite{ParamILS-JAIR}; indeed, we recover that method when the slack factor is 1. We allow for slack factors greater than 1 because they can improve model fits, albeit at the expense of more costly data acquisition.}
%
Figure \ref{fig:smbo-example} visualizes the first two steps of the resulting Bayesian optimization procedure for the minimization of a given blackbox function, starting from an initial Latin hypercube design.

We now return to algorithm configuration (AC), the problem motivating our research.
AC differs from standard problems attacked by Bayesian Optimization (BO) in some important ways: most importantly,
categorical input dimensions are common (due to algorithm parameters with finite, non-ordered domains);
inputs tend to be high dimensional; the optimization objective is a \emph{marginal} over instances (in the BO literature, this particular problem has, \eg{}, been addressed by~\cite{WilSanNot00,Groot:2010:BMC:1860967.1861017});
the objective varies exponentially (good settings perform orders of magnitude better than poor ones),
and the overhead of fitting and using models has to be taken into account since it is part of the time budget available for AC.
%
%
The model-based AC method \smac{} addresses these issues, including several modifications of standard BO methods to achieve state-of-the-art performance for AC (for details, see~\cite{HHLM10,HutHooLey11}). Here, we improve \smac{} further by setting censoring thresholds as described above and building models under the resulting censored data as described in Section \ref{sec:censored-regression}.
%

We compared our modified version of \smac{} to the original version on a range of challenging real-world configuration scenarios: optimizing the 76 parameters of the commercial mixed integer solver \cplex{} on five different sets of problem instances (obtained from \cite{HutHooLey10-mipconfig}), and the 26 parameters of the industrial SAT solver \spear{} on two sets of problem instances from formal verification (obtained from \cite{HutBabHooHu07}). Each \smac{} run was allowed 2 days and the maximum censoring time for each \cplex{}/\spear{} run was 10\,000 seconds.
Algorithm configuration scenarios with such high maximum runtimes have been identified as a challenge for \smac{} \cite{HutHooLey11},
and we demonstrate here that our adaptive censoring technique substantially improved its performance for these scenarios.

\NoCaptionOfAlgo
{
\DontPrintSemicolon
\small
\begin{algorithm}[tb]
\setcounter{AlgoLine}{0}
\caption[]{\AlTitleFnt{Algorithm \ref{alg:SMBO_general}: Sequential Model-Based Optimization (SMBO)}\\
{\footnotesize $\mathbf{R}$ keeps track of all target algorithm runs performed so far and their performances (\ie{}, SMBO's training data
$\{([\vtheta_1, \bm{x}_1], o_1), \dots, ([\vtheta_n, \bm{x}_n], o_n)\}$), $\mathcal{M}$ is SMBO's model, 
$\vec{\vTheta}_{new}$ is a list of promising configurations, and $t_{fit}$ and $t_{select}$ are the runtimes required to fit the model and select
configurations, respectively.
}}
\SetKwInOut{Input}{Input}
\SetKwInOut{Output}{Output}
\label{alg:SMBO_general}
\Input{Target algorithm $A$ with parameter configuration space $\vTheta$; instance set $\Pi$; cost metric $\hat{c}$}
\Output{Optimized (incumbent) parameter configuration, $\vtheta_{inc}$}

[$\mathbf{R}$, $\vtheta_{inc}$] $\leftarrow$ \emph{Initialize}($\vTheta$, $\Pi$)\;

\Repeat{total time budget for configuration exhausted}
{
$[\mathcal{M}, t_{fit}] \leftarrow$ \emph{FitModel}($\mathbf{R}$)\;
$[\vec{\vTheta}_{new}, t_{select}] \leftarrow $ \emph{SelectConfigurations}($\mathcal{M}$, $\vtheta_{inc}$, $\vTheta$)\;
$[\mathbf{R}, \vtheta_{inc}] \leftarrow$ \emph{Intensify}($\vec{\vTheta}_{new}$, $\vtheta_{inc}$, $\calM$, $\mathbf{R}$, $t_{fit} + t_{select}$, $\Pi$, $\hat{c}$)\;
}
\Return $\vtheta_{inc}$\;
\end{algorithm}
}
\RestoreCaptionOfAlgo

\begin{figure}[tbp]
  \begin{center}
				 \includegraphics[width=1\columnwidth]{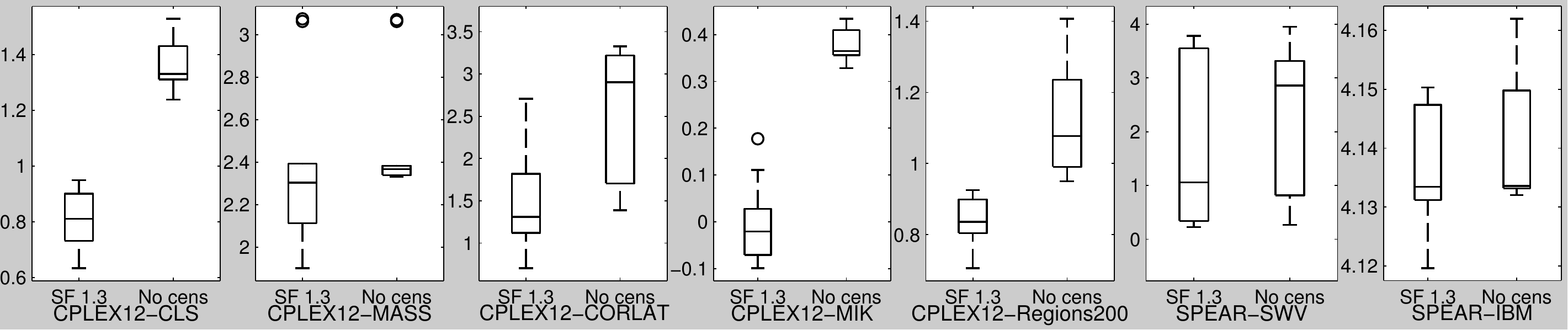}
    \small
    \caption{\small{}Visual comparison of \smac{}'s performance with and without censoring; note that the y-axis is runtime on a $log_{10}$ scale (lower is better).
    To avoid clutter, we only show capping with slack factor 1.3 (``SF 1.3'').
We performed 10 independent runs of each configuration procedure C and show boxplots of the test performances
for the resulting 10 final configurations (mean runtimes of \cplex{}/\spear{} across the test instances).     \label{fig:configuration-results}}
  \end{center}
\end{figure}

\begin{table}[tbp]
 \begin{center}
   {\small
   \begin{tabular}{c|c|cccccc}
     \hline
     \raisebox{-2mm}[1ex][1ex]{\textbf{Scenario}} & \raisebox{-2mm}[1ex][1ex]{\textbf{Unit}} & \multicolumn{6}{c}{\textbf{Median of mean runtimes on test set}}\\
     ~ & ~ & \textbf{SF 1} & \textbf{SF 1.1} & \textbf{SF 1.3} & \textbf{SF 1.5} & \textbf{SF 2} & \textbf{No censoring}\\
     \hline
CPLEX12-CLS & [$\cdot 10^0 s$] & $\bm{5.27}$ &$\bm{6.21}$ &$\bm{6.47}$ &     $8.3$  &     $6.66$  &     $21.4$  \\
CPLEX12-MASS & [$\cdot 10^2 s$] & $\bm{6.39}$ &$\bm{1.94}$ &$\bm{2.02}$ &$\bm{1.94}$ &$\bm{1.97}$ &$\bm{2.33}$ \\
CPLEX12-CORLAT & [$\cdot 10^0 s$] & $\bm{17.6}$ &$\bm{9.52}$ &$\bm{20.5}$ &$\bm{15.4}$ &$\bm{16.9}$ &     $826$  \\
CPLEX12-MIK & [$\cdot 10^{-1} s$] & $\bm{8.88}$ &$\bm{9.3}$ &$\bm{9.54}$ &$\bm{9.45}$ &$\bm{9.86}$ &     $23.9$  \\
CPLEX12-Regions200 & [$\cdot 10^0 s$] & $\bm{6.93}$ &$\bm{6.65}$ &$\bm{6.85}$ &$\bm{7.21}$ &$\bm{8.07}$ &     $12$  \\
     \hline
SPEAR-SWV & [$\cdot 10^0 s$] & $\bm{67.2}$ &$\bm{521}$ &$\bm{8.15}$ &$\bm{7.78}$ &$\bm{290}$ &$\bm{1030}$ \\
SPEAR-IBM & [$\cdot 10^4 s$] & $\bm{1.36}$ &$\bm{1.36}$ &$\bm{1.36}$ &$\bm{1.36}$ &$\bm{1.36}$ &$\bm{1.36}$ \\
     \hline
   \end{tabular}
   \vspace*{0.05cm}
   \caption{\small{}Comparison of \smac{} without and with censoring (using several slack factors, ``SF'').
   For each configurator and scenario, we report median test performance (defined as in
   Fig.~\ref{fig:configuration-results}; lower is better).
   We bold-faced entries for configurators that are not significantly worse than the best configurator for the respective scenario, based on a Mann-Whitney U test
   (note that the bold-facing of ``No censoring'' for SPEAR-SWV is not a typo: due to the very large variation for SPEAR-SWV visible in Figure \ref{fig:configuration-results} the Null hypothesis was not rejected).
   	\label{tab:configuration-results}}
		}
 \end{center}
 \vspace*{-0.15cm}
\end{table}

We performed 10 configuration runs for each of 7 problem domains and each of 6 versions of \smac{} (no censoring, and censoring with 5 different slack factors).
At the end of each configuration run, we recorded \smac{}'s best found configuration and computed the run's test performance as that configuration's mean runtime on a test set of instances disjoint from the training set, but sampled from the same distribution.

Figure \ref{fig:configuration-results} and Table \ref{tab:configuration-results} show that our modified version of \smac{} with censoring substantially outperformed the original \smac{} version without capping.
Our modified version gave better results in all 7 cases (with statistical significance achieved in 4 of these), with the improvements in median test performance reaching up to a factor of 126 (SPEAR-SWV).


\section{Conclusion}

We have demonstrated that censored data can be integrated effectively into Bayesian optimization (BO).
We proposed a simple EM algorithm for handling censored data in random forests and
adaptively selected censoring thresholds for new data points at small multiples above the best seen function values.
In an application to the problem of algorithm configuration, we achieved substantial speedups of the
state-of-the-art procedure \smac{}.
In future work, we would like to apply censoring in BO with Gaussian processes,
actively select the censoring threshold to yield the most information per time spent,
and evaluate the effectiveness of BO with censoring in further domains.



\subsubsection*{Acknowledgments}
We would like to thank our summer coop student Jonathan Shen for many useful discussions and for assistance in cleaning up the Matlab source code of \smac{} used for our experiments. Thanks also to Steve Ramage for porting \smac{} to Java afterwards and improving its usability, as well as for feedback on an earlier version of this report.

\bibliographystyle{plain}
\footnotesize{
\bibliography{frankbib}
}
\end{document}